\documentclass[conference]{IEEEtran}  %

\IEEEoverridecommandlockouts                              %

\usepackage{url}
\usepackage{cite}
\usepackage[pdftex]{graphicx}
\usepackage[caption=false,font=footnotesize]{subfig}
\usepackage{etoolbox} %
\usepackage{graphicx}
\usepackage{comment}
\usepackage{tabularx}
\usepackage{multirow}
\usepackage{booktabs}      %
\usepackage[table]{xcolor} %
\usepackage[nopatch=eqnum]{microtype}
\let\labelindent\relax
\usepackage{enumitem}
\usepackage{ifthen}

\def\CC{{C\nolinebreak[4]\hspace{-.05em}\raisebox{.4ex}{\tiny\bf ++}}}

\newbool{anonymous}
\setbool{anonymous}{false} %

\def\secref#1{Sec.~\ref{#1}}
\def\figref#1{Fig.~\ref{#1}}
\def\tabref#1{Tab.~\ref{#1}}
\def\eqref#1{Eq.~(\ref{#1})}

\usepackage{xspace}
\makeatletter
\DeclareRobustCommand\onedot{\futurelet\@let@token\@onedot}
\def\@onedot{\ifx\@let@token.\else.\null\fi\xspace}

\def\eg{\emph{e.g}\onedot} 
\def\ie{\emph{i.e}\onedot}

\makeatother

\begin{document}

\title{\LARGE \bf
Coral Grow-out Robotic Assessment System (CGRAS): Scaling Coral Recruit Monitoring Through Robotics and Computer Vision
}

\ifbool{anonymous}{
\author{Anonymous Submission}
}{
\author{Dorian Tsai$^{1}$, Scarlett Raine$^{1}$, Emilio Olivastri$^{1}$, Riki Lamont$^{1,2}$, Andrew Lui$^{1}$, Timothy Morris$^{1}$, Joshua Esplin$^{1}$,\\ Christopher A Brunner$^{2,3}$, F. Mikaela Nordborg$^{2}$, Reginald Wardleworth$^{1}$, Garima Samvedi$^{1}$, Karen Jackel$^{1}$,\\ Matthew Dunbabin$^{1}$, Tobias Fischer$^{1}$, Andrea Severati$^{2,3}$%
\thanks{$^{1}$Authors are with the Queensland University of Technology (QUT), Brisbane, Australia. Corresponding author: D.~Tsai, {\tt\small dy.tsai@qut.edu.au}}
\thanks{$^{2}$Authors are with the Australian Institute of Marine Science (AIMS), Townsville, Australia. R.~Lamont's contribution was conducted while affiliated with QUT.}%
\thanks{$^{3}$Authors are with the National Sea Simulator at AIMS.}
\thanks{The authors acknowledge the Traditional Owners of the land and sea Country where this research was performed, the Manbarra, Wulgurukaba and Bindal peoples. The adult corals and coral spawn used in this study were collected from the traditional sea Country of the Manbarra and Bindal peoples and were used with their Free Prior and Informed Consent. The authors are grateful for the granting of consent, acknowledge the place of Traditional Owners as the first scientists and custodians of their lands and pay our respects to Elders, past, present and emerging. 
We thank J. Terry, A. Goni, and A. Jones (QUT) for early contributions. 
We thank the QUT Research Engineering Facilities, Design and Fabrication Services and eResearch teams, and the AIMS Technology Transformation, Coral Aquaculture \& Deployment, and National Sea Simulator teams for their collaborative design, engineering, manufacturing, computational and coral husbandry support.
This research was supported by the Reef Restoration and Adaptation Program (RRAP) which is funded by a partnership between the Australian Government's Reef Trust and the Great Barrier Reef Foundation. 
D.T., S.R., E.O. and T.F. acknowledge support from the QUT Centre for Robotics; T.F. acknowledges Australian Research Council Discovery Early Career Researcher Award Fellowship DE240100149.%
}}}

\def\CC{{C\nolinebreak[4]\hspace{-.05em}\raisebox{.4ex}{\small\bf ++}}}

\maketitle
\thispagestyle{empty}
\pagestyle{empty}

\begin{abstract}

Climate change is the largest threat to coral reefs, with increasing global impacts accelerating the need for scalable reef restoration technologies. Large-scale reef restoration depends on the mass production of corals, such as through coral aquaculture. Coral seeding with recruits grown in aquaculture facilities is a feasible restoration approach, but effective production requires consistent, high-frequency monitoring of tens of thousands of macroscopic ({$\approx$}0.5–2mm diameter) recruits, making conventional manual assessment prohibitively labor-intensive. To address this monitoring bottleneck, we introduce the Coral Grow-out Robotic Assessment System (CGRAS) which combines robotic imaging and computer vision to automate data acquisition, perform multi-species detection and counting of corals, and evaluate coral health. CGRAS automatically extracts coral growth, survival and spatial distribution metrics, with the aim of providing timely feedback to operators for optimizing production, grow-out and deployment workflow processes. We demonstrate CGRAS in a large aquaculture facility on standardized coral settlement tiles, reducing the time and labor costs by a factor of 9.6 as compared to manual monitoring, whilst achieving 96.4\% agreement for \textit{Acropora kenti} corals relative to expert counts. 
\end{abstract}

\section{Introduction}
\label{sec:intro}

Coral reefs are under threat from both local and global anthropogenic stressors, with mass coral bleaching events increasing in severity and frequency over the past two decades~\cite{hughes2018spatial}. Bleaching is a stress response in which corals lose their symbiotic algae, their primary energy source, and is most commonly due to heat stress driven by global climate change~\cite{hughes2017coral}. If these conditions persist, bleaching can ultimately lead to coral death~\cite{glynn1993coral}. Without effective intervention~\cite{de201227}, substantial reef loss is projected by 2050, with significant consequences for marine life and the ecosystem services reefs provide~\cite{hoegh2007coral}. Decisive action on climate change remains essential to preserve coral reef ecosystems, but additional management strategies such as coral reef restoration are increasingly necessary to maintain reef resilience.

\ifbool{anonymous}{One of the most promising approaches for large-scale reef restoration is \textit{coral seeding}}{The Reef Restoration and Adaptation Program (RRAP), the world's largest coordinated reef restoration initiative, recognized \textit{coral seeding} as a highly scalable and effective strategy for reef restoration}~\cite{ramsby2026developing}. In this process, the sexual reproduction of broadcast spawning corals is leveraged to produce millions of corals that can be grown out in nurseries and subsequently transplanted onto degraded reefs~\cite{severati2024autospawner, tsai2026cslics, ramsby2026coral}. This approach also facilitates the use of temperature-resistant parent corals and maintains high genetic diversity of the offspring, thereby improving adaptive potential while generating the numbers of corals needed to support large-scale restoration interventions~\cite{severati2024autospawner}.

\ifbool{anonymous}{
\begin{figure}[t]
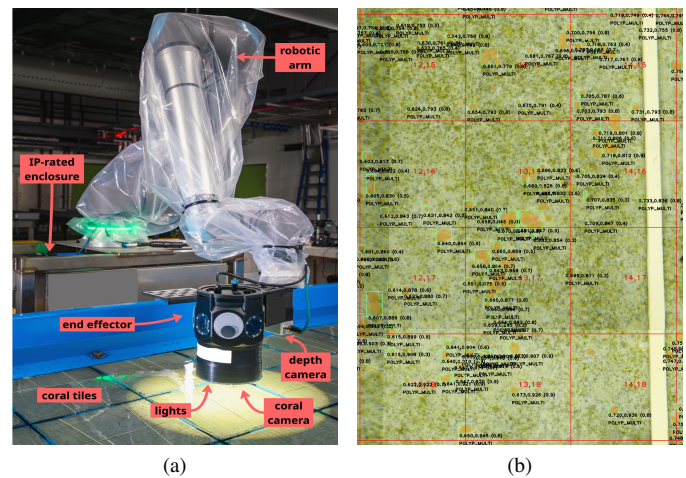

    \centering
    \subfloat[]{\includegraphics[height=0.65\columnwidth, width=0.485\columnwidth]{figures/CGRAS_Hardware_Overview_Main_Annotated_Anon.pdf}} \hfill
    \subfloat[]{\includegraphics[height=0.65\columnwidth, width=0.485\columnwidth]{figures/CGRAS_example_detections_closeup.png}}
    \caption{Coral Grow-out Robotic Assessment System (CGRAS). (a) Image capture system. (b) Magnified example of coral recruit detections, where the red lines denote the $14\times14$mm tabs that the $280\times280$mm coral settlement tiles are broken into prior to deployment.}
    \label{fig:cgras}
    \vspace{-1.0em}
\end{figure}
}{
\begin{figure}[t]
    \centering
    \subfloat[]{\includegraphics[height=0.65\columnwidth, width=0.485\columnwidth]{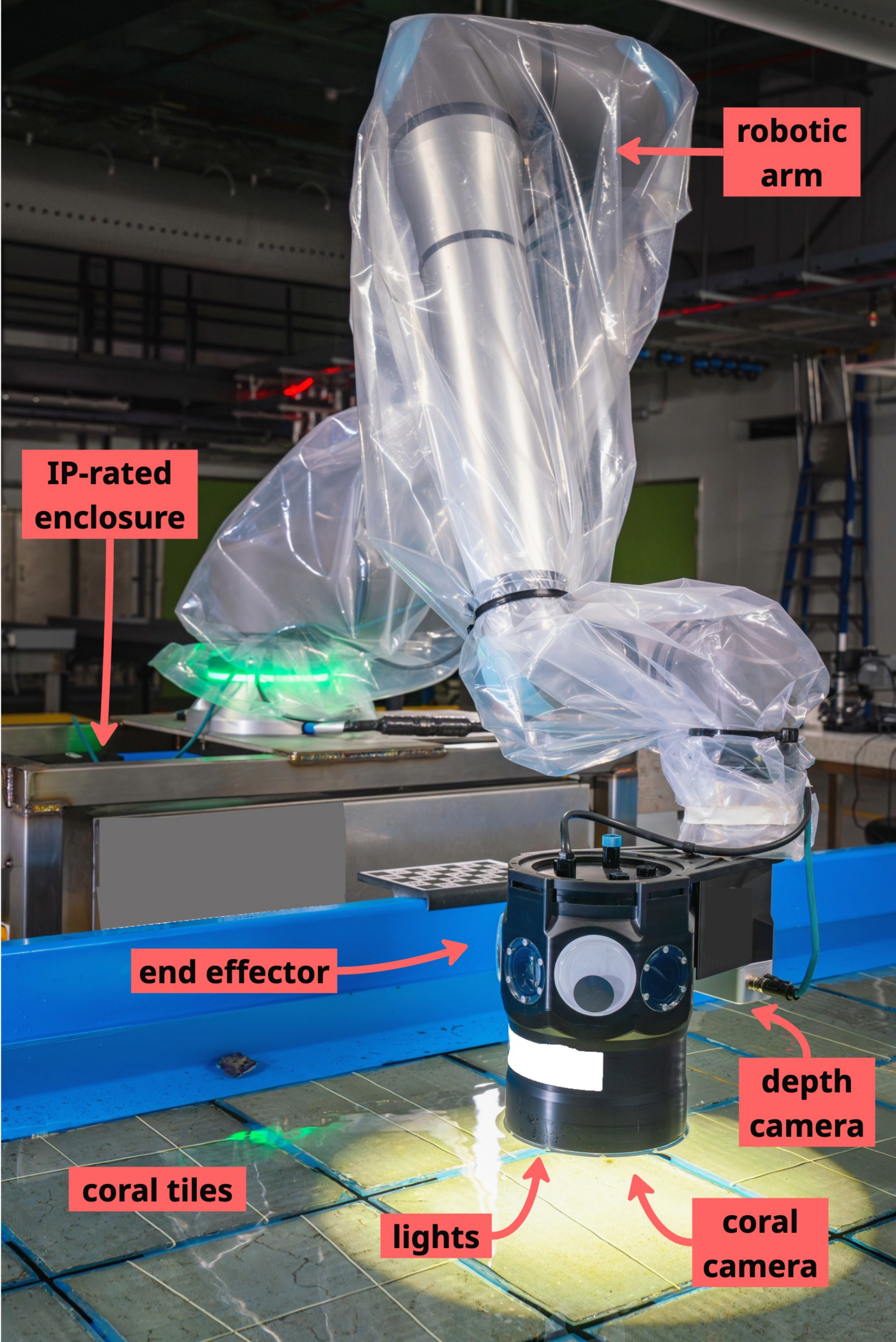}} \hfill
    \subfloat[]{\includegraphics[height=0.65\columnwidth, width=0.485\columnwidth]{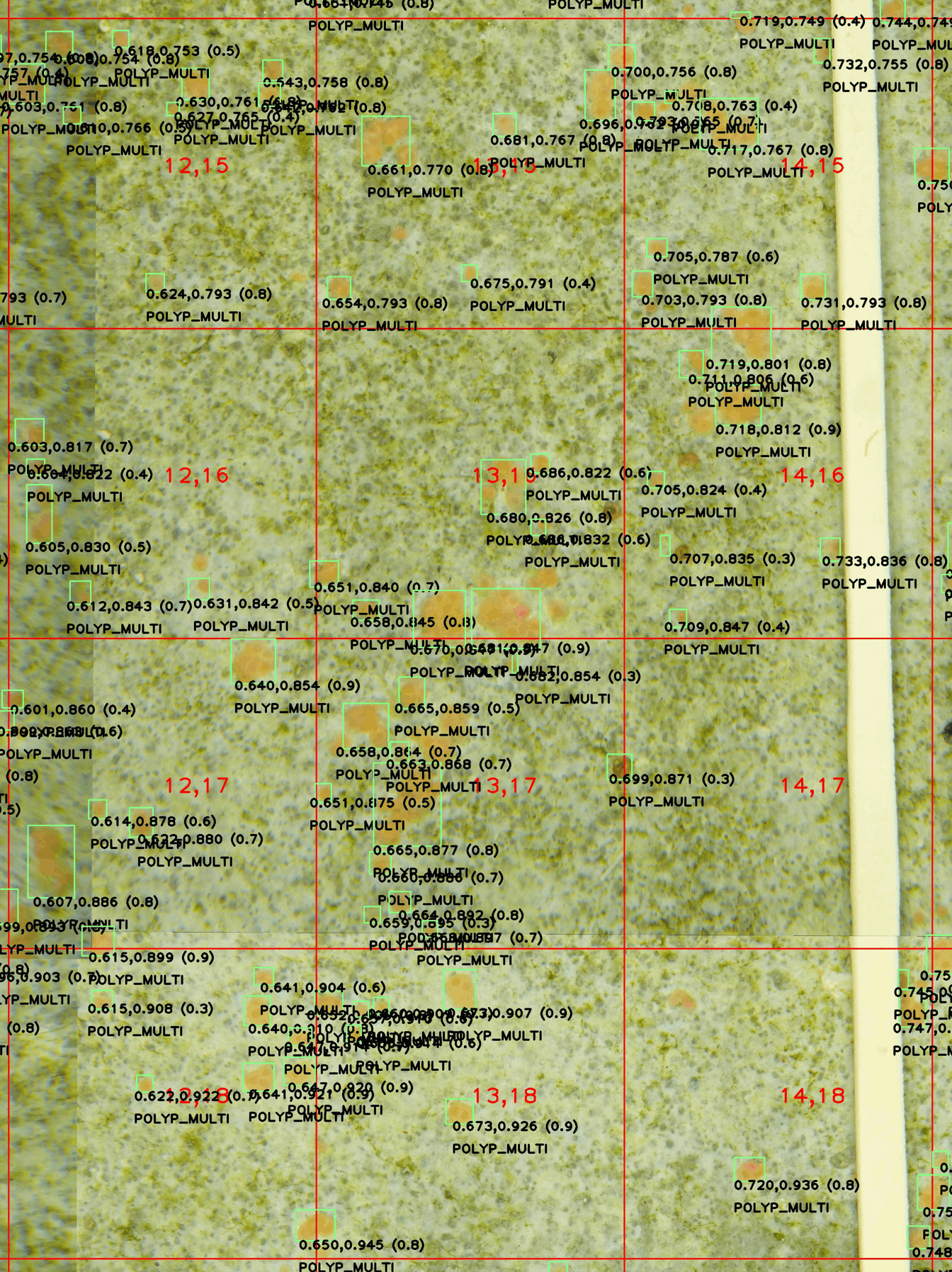}}
    \caption{Coral Grow-out Robotic Assessment System (CGRAS). (a) Image capture system. (b) Magnified example of coral recruit detections, where the red lines denote the $14\times14$mm tabs that the  $280\times280$mm coral settlement tiles are broken into prior to deployment.}
    \label{fig:cgras}
    \vspace{-1.0em}
\end{figure}
}

The coral \textit{grow-out} phase begins following fertilization of coral spawn: developing larvae settle onto dedicated substrate, \eg, concrete tiles and mature into \textit{recruits}, which are then cultivated over approximately 12 weeks under controlled conditions. This phase is highly sensitive to environmental factors~\cite{edmunds2023coral} and presents substantial challenges to large-scale production and restoration. Grow-out tanks are designed to closely replicate favorable ocean conditions~\cite{ramsby2026coral}, yet since the drivers of successful cultivation are not yet fully understood, marine scientists must perform frequent inspections to tune tank parameters and maximize cultivation success.

Currently, assessing the state of a single $280\times280$mm concrete tile requires three different steps: transporting it to a stereoscope station, manually annotating and counting coral recruits, and returning it to the tank. An accurate assessment of a tile typically requires 1–3 hours of stereoscope work by a trained expert. In addition to the significant labor required, the manual handling required for tile assessments may cause additional stress to corals, potentially affecting survival. Automation of monitoring is therefore essential to meet future coral production and deployment targets.

Automating this task is complicated by two compounding factors: first, coral recruits are millimeter-scale organisms that must be resolved at microscope-comparable resolution, while a robotic system operates at a meter-scale reach across multiple tank arrays. Second, all hardware must continuously operate in a humid, saline, marine environment without compromising image precision or calibration accuracy. 

\begin{figure*}[t]
\centering
\includegraphics[width=\linewidth]{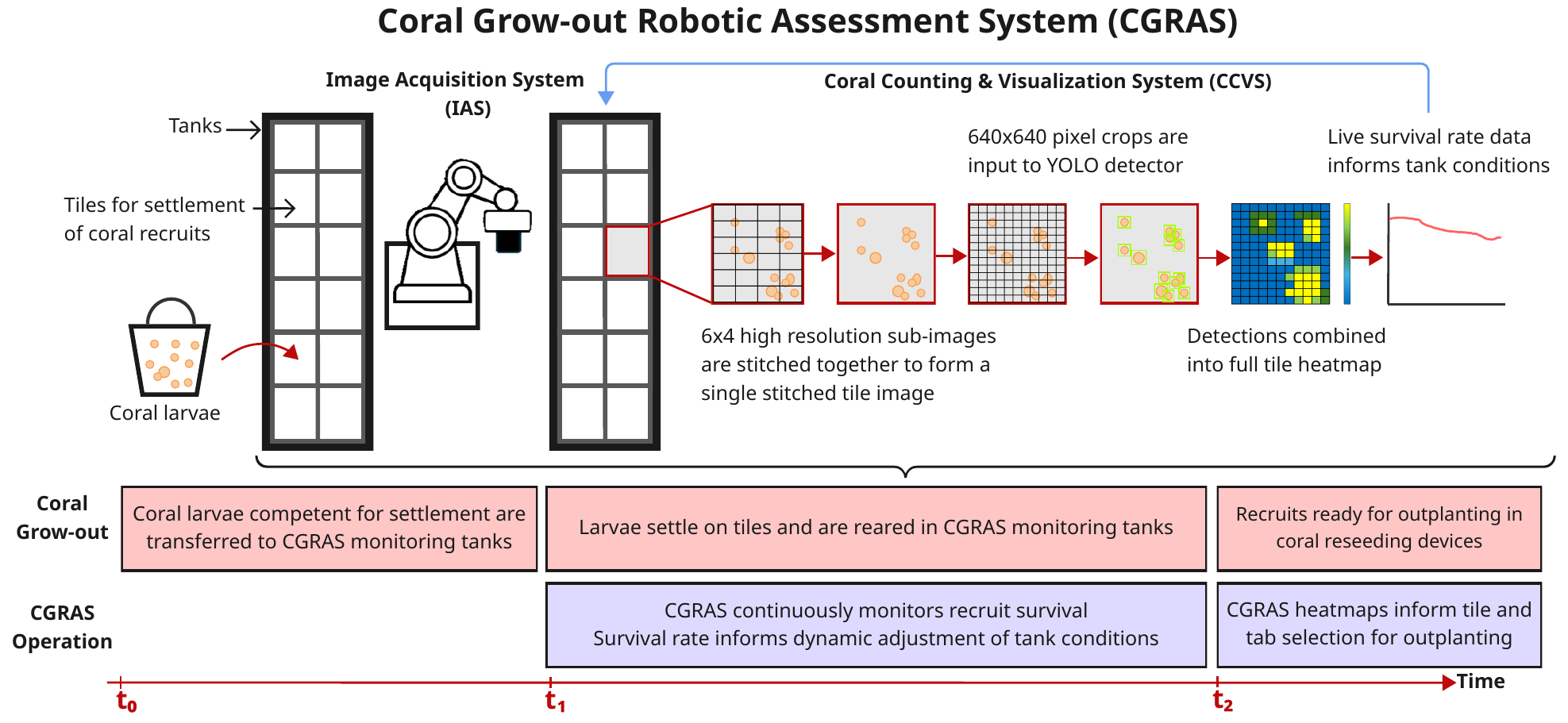}
\vspace{-7.0mm}
\caption{Coral grow-out starts when competent coral larvae settled on provided concrete tiles (t\textsubscript{0}). During operation, CGRAS performs continuous monitoring of coral survival (t\textsubscript{1}-t\textsubscript{2}), ranging from 2 to 12 weeks, enabling real-time optimization of tank conditions. When coral recruits have reached an appropriate growth stage for reseeding onto the reef (t\textsubscript{2}), the CGRAS heatmaps are used to select the specific tiles and tabs for outplanting.}
\label{fig:overview}
\vspace{-5.0mm}
\end{figure*}

To address these challenges, we introduce the Coral Grow-out Robotic Assessment System (CGRAS;~\figref{fig:cgras}), which enables timely extraction of large-scale coral growth metrics unattainable through manual monitoring methods by combining two bespoke subsystems:
\emph{The Image Acquisition System (IAS)}, a robotic imaging system that automates collection of ultra-high-resolution coral datasets through non-invasive repeatable image acquisition at microscopic scales in marine experimental and aquaculture facilities, and 
\emph{The Coral Counting \& Visualization System (CCVS)}, a data processing pipeline that efficiently processes high-resolution coral imagery to estimate coral counts, assess coral health, and analyze spatial distribution over time, quickly producing rich datasets for advancing coral research and real-time decision making.

\section{Related Work}
\label{sec:related}

CGRAS lies at the intersection of three research fields: automated analysis of early-life-stage coral imagery, imaging hardware for benthic monitoring, and robotic and learning-based systems for reef-scale assessment.

\subsection{Automated Assessment of Early Coral Life Stages}
Computer vision has been used to automatically assess coral growth. For example, image-based methods have automated assessment from the larval stage onwards. Large-particle flow cytometry enables rapid counting and spectral classification of free-swimming larvae~\cite{randall2020rapid}, and machine learning classifies larval fitness from back-lit microscope images on glass slides~\cite{macadam2021machine}, establishing that automated assessment is feasible under controlled conditions. Most relevant, a hierarchical YOLO-SAM pipeline segments and tracks coral recruits from time-series microscopy, exceeding 95\% mean Intersection over Union (IoU) while notably reducing analysis time~\cite{zhao2026hierarchical}. These methods assume imagery already exists: each relies on samples or tiles imaged individually under a microscope or stereoscope. No prior approaches perform automated image acquisition for tiles in a coral production facility.

\subsection{Imaging Platforms for Benthic Monitoring}
Acquiring recruit-scale imagery in-situ has motivated dedicated optical platforms. Diver-deployable underwater microscopes resolve micron-scale benthic structure~\cite{mullen2016underwater} and have been extended with focal scanning and photosynthetic-efficiency measurement~\cite{ben2025benthic} and with extended depth of field via focus stacking~\cite{shahani2021design}. These instruments excel at long-term, single-point deployments but are not built for batch imaging of many substrates. Commercial aquatic imaging systems~\cite{cpics, pi10, ISIIS-DPI} instead target free-floating particles or flow-through fish aquaculture, and lack the combined field-of-view and substrate-level resolution that post-settlement recruits require.

\subsection{Robotic and Learning-based Reef Monitoring}
At reef scale, robotics and deep learning are already used together for monitoring and intervention. Uncrewed surface vehicles support reef mapping and larval dispersal~\cite{dunbabin2020uncrewed, mou2022reconfigurablerobots}, while AI-based methods have been demonstrated for guiding the robotic deployment of coral reseeding devices~\cite{raine2025ai}. Deep learning has also been used to estimate coral cover from photographic transects~\cite{gonzalez2020monitoring}, underwater video~\cite{li2024deep}, and point-labeled benthic imagery~\cite{raine2022superpixels}, while patch-based detection on towed-camera imagery contends with variable illumination, biofouling, and rare taxa~\cite{trotter2025automated}. All of these operate at spatial resolutions orders of magnitude coarser than an individual recruit. Within aquaculture facilities, the Coral Spawn \& Larvae Imaging Camera System (CSLICS)~\cite{tsai2026cslics} automates monitoring of spawn and free-swimming larvae, saving thousands of hours per season, but this system cannot be used to monitor recruits following settlement. Post-settlement recruits are more challenging to monitor than larvae as they are attached to complex surfaces, spatially heterogeneous, and morphologically dynamic throughout the grow-out phase~\cite{edmunds2023coral}.

CGRAS closes this gap by pairing a robotic arm and high-resolution underwater imaging system with a purpose-built detection pipeline for automated, high-throughput monitoring of coral recruits in aquaculture grow-out.

\ifbool{anonymous}{
\begin{figure}
    \centering
    \includegraphics[width=1.0\columnwidth]
    {figures/CGRAS_Hardware_Overview_Annotated_Anon.pdf}
    \caption{The IAS captures two coral imaging tanks in a single operation. Calibration tags establish the spatial relationship between the robot arm base and the imaging tanks, allowing accurate localization of coral tile positions within tanks.}
    \label{fig:hardware3}
    \vspace*{-0.5cm}
\end{figure}
}{
\begin{figure}
    \centering
    \includegraphics[width=1.0\columnwidth]
    {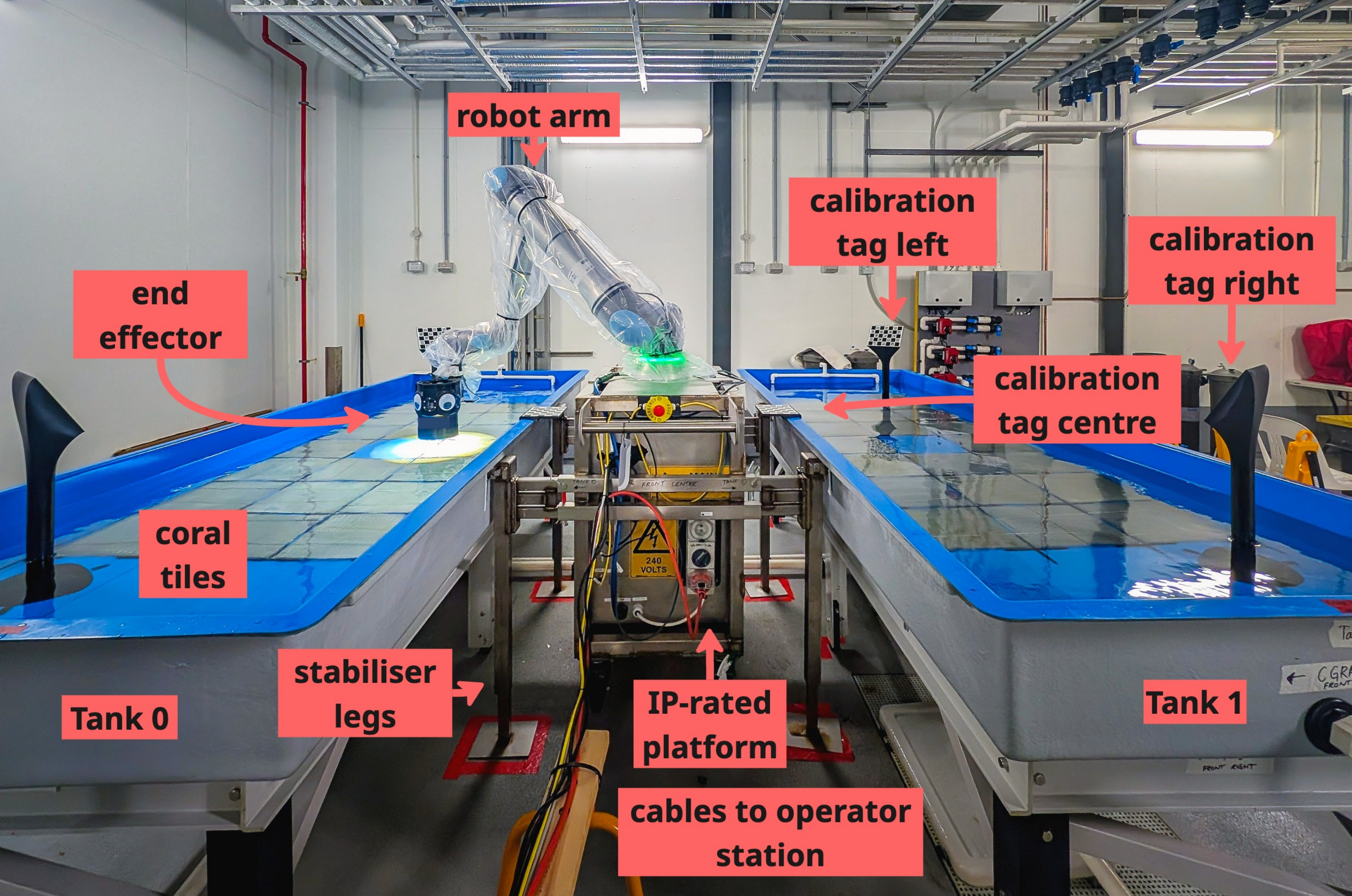}
    \caption{The IAS captures two coral imaging tanks in a single operation. Calibration tags establish the spatial relationship between the robot arm base and the imaging tanks, allowing accurate localization of coral tile positions within tanks.}
    \label{fig:hardware3}
    \vspace*{-0.5cm}
\end{figure}
}

\section{Method}
\label{sec:method}
CGRAS is a scalable, reliable, and cost-effective solution for coral health monitoring during the recruit grow-out phase of production. CGRAS consists of two primary subsystems: the Image Acquisition System (IAS) is a waterproof robotic imaging system that captures an array of high-resolution microscopic images across an entire coral growth tile. The Coral Counting \& Visualization System (CCVS) is a data processing pipeline that stitches these images into a unified tile-level representation. The CCVS also automatically detects and annotates alive corals, and generates heat maps that visualize the spatial distribution of healthy coral recruits (\figref{fig:overview}).

\subsection{Preliminaries: the Coral Grow-out Process}
\label{subsec:method-process}

This section provides an overview of the coral conservation aquaculture workflow. Coral colonies are collected from the reef and brought to \ifbool{anonymous}{a research aquaculture facility}{the National Sea Simulator (SeaSim) at the Australian Institute of Marine Science (AIMS)} where they are maintained in tanks under conditions similar to their reef of origin until spawning occurs. During spawning, egg–sperm bundles are released and then automatically harvested. Fertilized eggs are transferred to larval culture tanks, where they develop into free-swimming larvae over the following week~\cite{ramsby2026coral} and are monitored by automated camera systems~\cite{tsai2026cslics,severati2024autospawner}. Once mature, larvae are settled onto preconditioned concrete tiles (280$\times$280 mm), manufactured as a $20\times20$ grid of tabs~\cite{ramsby2026coral}. 

Recruits are subsequently reared in dedicated tanks for up to 12 weeks during the coral grow-out stage, where tiles are continuously assessed for suitability for deployment based on coral size and density. Shortly before planned deployments, tiles are separated into individual tabs and further evaluated using geometric and coral density criteria. Suitable tabs with healthy corals are then fitted into coral seeding devices before being transported to the target reef and deployed on appropriate reef substrates~\cite{raine2025ai, ramsby2026coral}.

\subsection{Image Acquisition System (IAS)}
\label{subsec:method-image-acquisition}
The IAS is the first component of the CGRAS and is responsible for capturing images of grow-out tiles. It is composed of the following hardware components (\figref{fig:hardware3}):

\paragraph{Robot Arm} A Universal Robots UR20 robotic arm is mounted on top of a mobile platform to capture images of the tiles within the imaging tanks. The use of a robotic arm provides flexibility to accommodate future changes in aquaculture facility infrastructure while ensuring accurate and repeatable image acquisition. The UR20 offers sufficient reach to cover all required imaging locations and is straightforward to configure. As a collaborative robot, it is designed for safe operation alongside human operators, while its IP65 rating makes it suitable for splash-prone aquaculture facilities. To improve durability under corrosive conditions, the arm is enclosed in a transparent protective plastic sleeve. The system is controlled through the UR20 controller and the Robot Operating System (ROS) framework~\cite{quigley2009ros}.
    
\paragraph{End Effector} The end effector houses the scientific imaging camera, a depth camera, lights and a Raspberry Pi in a waterproof flat-port enclosure. The imaging camera is a Sony ILX-LR1 with a 50mm F2.8 macro lens, capturing $9,428 \times 6,309$ pixel sub-images. Illumination is provided by two series of four high-power white LEDs (Luminus MP-5050-250R-40-70), delivering 3,000 lumens at the imaging distance of ${\approx}14$cm from the housing. The LEDs are arranged in a ring-light configuration around the lens to ensure uniform scene illumination. A Raspberry Pi 4B running Ubuntu 20.04 LTS controls both the camera, via its SDK, and the lighting system, and prevents overheating failures and ensures stable operation. The Framos D435e IP66 industrial depth camera, used for hand-eye calibration, is mounted outside the enclosure.
    
\paragraph{Platform} The system is built on a welded stainless-steel mobile frame to resist corrosion and incorporates 100 kg of ballast for stability, along with emergency stop buttons and a safety light tower. Computing, networking, safety, and control hardware for robot operation and image processing are housed within an IP68-rated enclosure, which sits within the frame, providing protection against splashes, humidity, and saltwater exposure. Extendable stabilizer legs are used to level and secure the platform, preventing instability or tipping during operation. 
    
\paragraph{Imaging Tanks} The fiberglass tanks where the tiles are placed measure approximately $330 \times 130 \times 35$cm ($L \times W \times H$). Each tank accommodates up to 25 tiles, with the outermost positions reserved for calibration patterns. The current CGRAS setup consists of two tanks, with CGRAS positioned between them, enabling imaging of up to 50 tiles in a single session. 

\paragraph{RFID Tags} The coral tiles are secured in semi-permanent polymer or silicone molds that have a radio frequency identification (RFID) tag, enabling unique association between each imaged tile and its corresponding coral health metrics. Although the RFID range is substantially reduced in water, this limitation is beneficial as it minimizes cross-talk during the scanning of the tags.  

Next, we describe the complete procedure for collecting imagery. First, coral tiles are placed inside the imaging tanks by the operator and CGRAS is started. Next, a handheld RFID reader (\ie, Biomark HPRLite) is used to scan the tiles in a predefined, configurable sequence to associate coral health metrics with the corresponding imaged tiles. Hand-eye calibration is performed through the IAS interface using the depth camera. To robustly estimate the pose of the tank relative to the robot arm base, three calibration tags placed at predefined and marked locations are used (\figref{fig:hardware3}).

Once calibration is completed, the robot arm follows a predefined, user-configurable open-loop path to capture a $6\times 4$ grid of overlapping sub-images per tile, allowing the sub-images to be combined into a single stitched tile image. Finally, after a tank has been fully imaged, the acquired imagery is processed using the CCVS.

\subsection{Coral Counting \& Visualization System (CCVS)}
\label{subsec:method-counting}

The CCVS processes the data acquired by the IAS. It consists of software modules for data management, image stitching, tile localization within the stitched tile image, automated coral detection, coral health trend analysis, and generation of heatmaps describing the spatial distribution of corals across individual tiles. All outputs are accessible through the CCVS web interface, which supports visualization and data extraction.

Notably, the image stitching module combines each batch of 24 sub-images into a single tile image of approximately $28,000\times28,000$ pixels, and the tile localization module then determines the tile position within the stitched image, which is necessary for producing coral distribution heatmaps. The physical corners of the tiles are unreliable visual cues: variation in the manufacturing process can result in concrete spreading beyond the intended tile boundaries, partially or completely obscuring the true tile corners. We therefore designed black corner tile mold spacers with consistent geometry.  We detect these corners using a Random Forest-based classifier for image binarization and then perform template matching, providing a more reliable reference for tile stitching.

The coral detection module identifies alive corals within the stitched tile image. Since direct inference on the stitched image is computationally impractical and down-sampling would discard fine visual details essential for accurate detection, the module adopts a sliced inference strategy inspired by~\cite{akyon2022sahi}: the stitched image is divided into overlapping $640 \times 640$ pixel patches, which are processed individually before results are aggregated.

Each patch is processed independently using a YOLOv8n object detection model~\cite{ultralyticsyolov8} (refer to~\secref{subsec:model} for model training details). This model was selected for its lightweight design and high inference speed, which are essential given that a single tile image requires more than 3,000 inference passes. Overlapping detections are then merged to eliminate duplicate coral counts.

The analysis module processes the detected corals to generate temporal health trends and spatial heatmaps. Heatmaps partition the tile into a $20\times20$ grid of tabs, highlighting regions of coral growth across each tile. These outputs are accessible through the CCVS web-based interface, allowing operators to visualize growth trajectories and adjust tank conditions to maximize recruit survival.

\begin{figure}
    \centering
    \includegraphics[width=1.0\columnwidth]{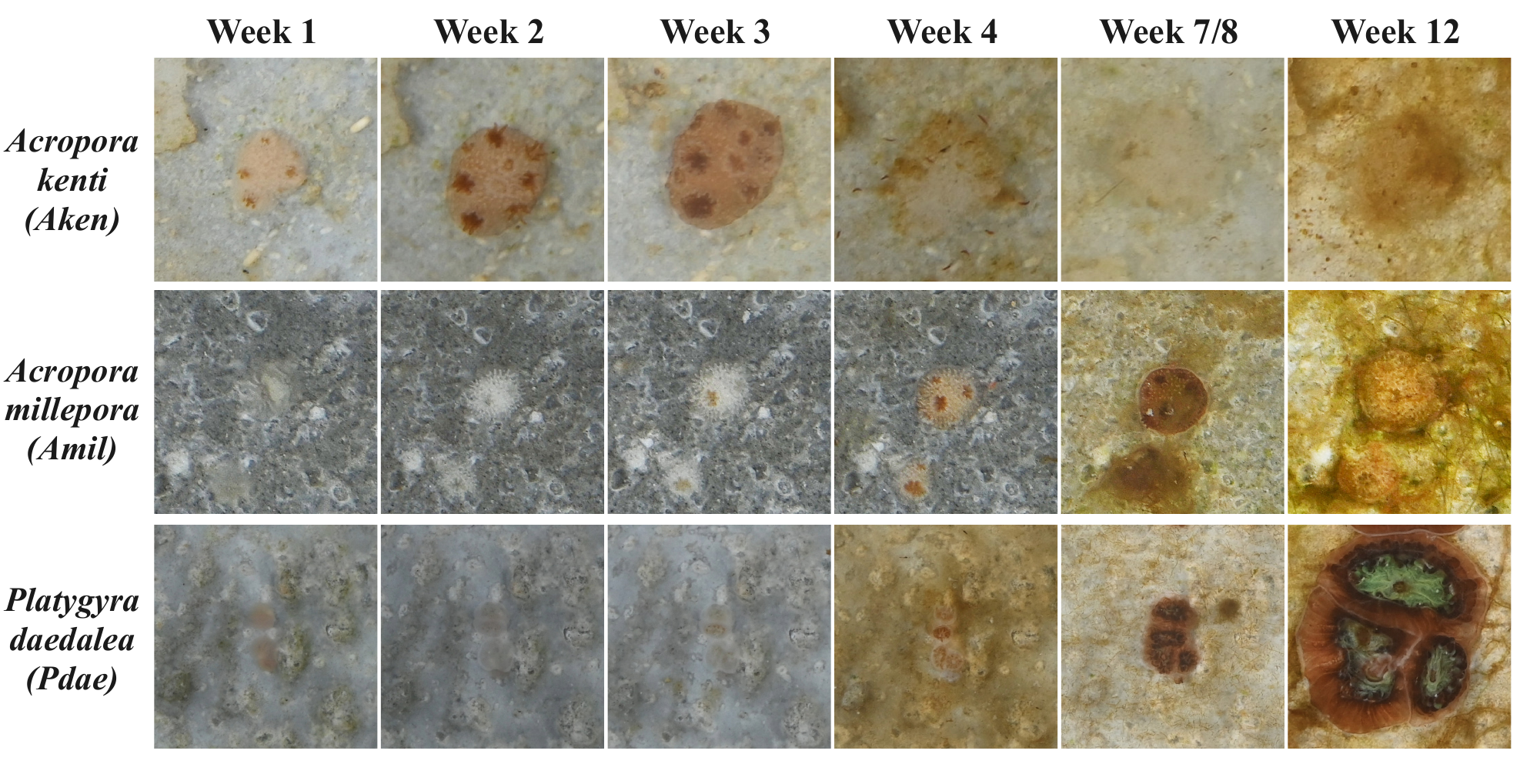}
    \caption{Examples of coral development for \textit{Aken}, \textit{Amil} and \textit{Pdae} coral recruits. During Weeks 0--1, recently settled corals begin forming skeletal structures that are only faintly visible. Beyond Week 1, skeletal development becomes more pronounced,  while the central polyp becomes progressively darker due to the increasing abundance of symbiotic \textit{Symbiodiniaceae} algae within the coral tissue.  Coral mortality is also seen after Week 4 for \textit{Aken}, and Week 12 for \textit{Amil}, where soft tissue degradation is followed by overgrowth of algae and other benthic organisms on the exposed skeleton. \textit{Pdae} are smaller, and largely transparent until Week 4 when they grow in size and visually differ from \textit{Aken} and \textit{Amil}.}
    \label{fig:coral_examples}
    \vspace{-1.0em}
\end{figure}

\section{Experimental Setup}
\label{sec:experiments}

CGRAS was deployed at \ifbool{anonymous}{a research facility}{SeaSim} where corals are grown under controlled conditions. This study utilized datasets collected during the 2024 and 2025 late spring/early summer mass spawning events and their subsequent grow-out periods.

\subsection{Data Collection}
\label{sec:datacollection}
Coral species used for data collection were selected based on the availability of gravid colonies, spawn quality, the number of parent colonies, and suitability for use in future deployment activities on the reef. The primary species included in the dataset were \textit{Acropora kenti} (\textit{Aken}) and \textit{Acropora millepora} (\textit{Amil}), both of which are well-studied, fast-growing corals of the genus \textit{Acroporidae}, commonly used in reef restoration research (Fig.~\ref{fig:coral_examples}). Additionally, \textit{Platygyra daedalea} (\textit{Pdae}) was included to represent a slower-growing coral with distinct visual morphology, as seen in~\figref{fig:coral_examples}. Corals were reared as described in Section~\ref{subsec:method-process} and imaged weekly with CGRAS during grow-out.

\subsection{Data Annotation}
\label{subsec:data}
Coral recruits were annotated as a single live-coral class for the analyses presented in this work. Annotations were generated using the Computer Vision Annotation Tool (CVAT)~\cite{boris_sekachev_2020_4009388}, where each coral instance was labeled using a bounding box.

Due to the substantial effort required to annotate high-resolution coral imagery, optimizing the annotation workflow was critical. Ten sub-images were initially manually annotated and partitioned into $640 \times 640$ pixel patches using the same sliced-image approach adopted during inference. These labeled patches were then used to train a preliminary coral detection model, which generated candidate annotations for previously unlabeled sub-images. The predicted annotations were manually reviewed and corrected in CVAT to produce accurate ground truth labels.
 
This bootstrapping strategy, inspired by~\cite{barth2019synthetic}, significantly reduced annotation effort. The first thirty manually labeled images required 31 minutes per sub-image on average by a trained expert. After introducing the model-assisted annotation workflow, the annotation time for the remaining 110 sub-images per coral species decreased to 6.5 minutes per sub-image.

\subsubsection{2024 Spawning Dataset} 
Data was collected for all three coral species (\ie, \textit{Aken}, \textit{Amil} and \textit{Pdae}) during the 2024 mass spawning event with images manually labeled to train the CGRAS coral detection network. The 2024 dataset consists of images captured for 18 tiles spanning the 12-week grow-out period. Each tile image was composed of 24 sub-images, resulting in a total of 3,024 sub-images per species. To create the training dataset, 140 sub-images per species were selected across the tiles and annotated with bounding boxes following the CVAT bootstrapping strategy.

\subsubsection{2025 Spawning Dataset} 
Data was collected for \textit{Aken} and \textit{Amil} corals during the 2025 mass spawning event with images manually labeled to train the CCVS coral detection network. The 2025 dataset consists of images captured for 99 tiles of \textit{Aken} and 83 tiles of \textit{Amil} during the first 4 weeks post-settlement. Each tile image was composed of 24 stitched sub-images, resulting in 2,376 sub-images for \textit{Aken} and 1,992 sub-images for \textit{Amil}. Due to limited expert annotation time, 10 tile images per species were manually annotated with bounding boxes in CVAT for model training.

\begin{table}[t]
  \centering
  \caption{Operational performance of the CGRAS system, evaluated during 2025 spawning season. See~\secref{subsec:results_imaging} for details.}
  \label{tab:cgras_operational}
  \footnotesize
  \setlength{\tabcolsep}{5pt}
  \renewcommand{\arraystretch}{1.2}
  \begin{tabular}{@{}llrr@{}}
    \toprule
    \textbf{Category} & \textbf{Metric} & \textbf{Value} & \textbf{Unit} \\
    \midrule
    \multirow{3}{*}{Acquisition: IAS}
      & Tiles imaged           & 183  & tiles \\ 
      & Non-recoverable errors & 1    & tile \\
      & Imaging time    & 14.6 & hours \\
 
    \midrule
    \multirow{3}{*}{Processing: CCVS}
      & Total tiles & 182  & tiles  \\
      & Successfully processed tiles  & 160 & tiles     \\
      & Processing time        & 7.6 & hours  \\
 
    \midrule
    \multirow{2}{*}{Full System: CGRAS}
      & Total time             & 22.2        & hours \\
      & Equivalent manual time      & 213.5       & hours \\
 
    \bottomrule
  \end{tabular}
  \vspace{-1.0em}
\end{table}

\subsection{Model Training and Inference}
\label{subsec:model}

For each dataset year (2024 and 2025), images for each coral species were treated as separate datasets and partitioned into training, validation, and test subsets using a 0.7/0.15/0.15 split. Training was conducted on an NVIDIA H100 GPU using COCO pre-trained weights until convergence or for a maximum of 1,000 epochs. Standard data augmentation was used during training to improve model robustness. 

During inference, the CCVS runs on the CGRAS onboard computer equipped with an Intel 12th Gen i7 12-core CPU and an NVIDIA RTX 3060 12GB GPU. 

\begin{figure}[t]
    \centering
    \subfloat[]{\includegraphics[width=0.48\columnwidth]{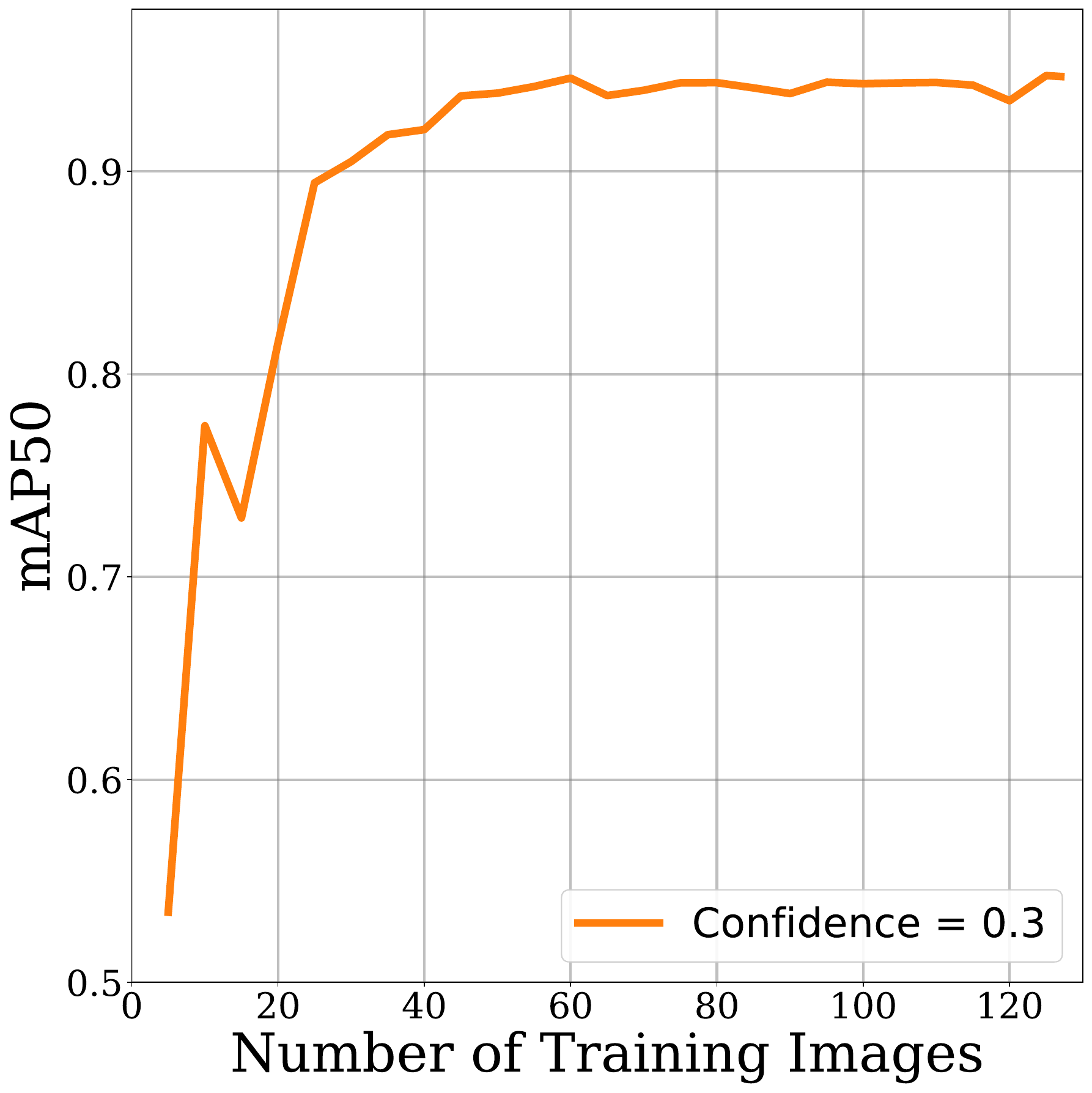}}\hfill
    \subfloat[]{\includegraphics[width=0.48\columnwidth]{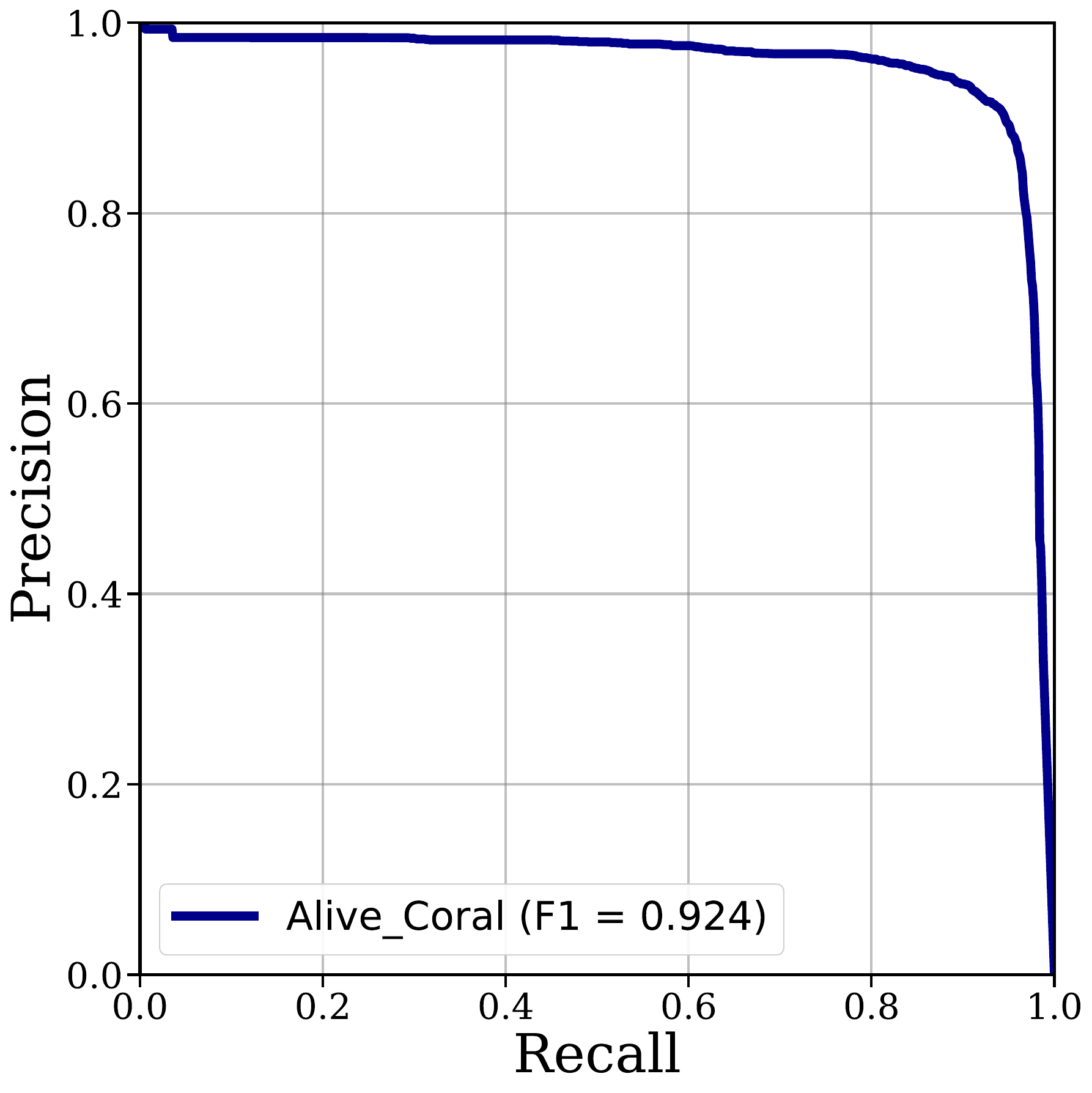}}
    \caption{Performance of CCVS object detection models for the coral species \textit{Aken}. (a) mAP50 as a function of the number of labeled training images, averaged across nine independent training runs with different random seeds and data shuffles (refer to~\secref{subsec:metrics} for metric defintions). Performance increased rapidly up to approximately 30 labeled images before plateauing, indicating that 30 images provide a practical balance between annotation effort and model performance in data-constrained settings. (b) Precision--Recall curve for the alive coral class of the final model, trained using all 130 labeled images. Although most gains were achieved with relatively few labeled images, the final model used for ML-assisted labeling was trained on the complete dataset to maximize performance.}
    \label{fig:detection_results}
    \vspace{-1.0em}
\end{figure}

\subsection{Evaluation Metrics}
\label{subsec:metrics}
The IAS is evaluated through operational metrics including: number of tiles successfully imaged, number of non-recoverable errors and imaging time in hours. The CCVS is also assessed in terms of its operational performance, with evaluation of the number of tiles successfully processed by the CCVS and the processing time. The overall time for CGRAS to perform an imaging session was also reported and compared against the equivalent manual time it would take a trained marine ecologist to manually produce the equivalent data.

For the detection component of the CCVS, performance was evaluated using the mean Average Precision at an Intersection over Union threshold of 0.5 (mAP50). When calculating mAP50, a confidence threshold of 0.3 was used.  

We evaluate the agreement between CGRAS coral counts and manual expert counts by calculating the difference between CGRAS and manual counts on a per-tab basis, and report the distribution of differences, the mean and the standard deviation.

\section{Results}
\label{sec:results}

In the following sections, we evaluate the performance of CGRAS in terms of its ability to reliably acquire tile imagery, detect and classify corals, assess the coral density and distribution on tiles, and track coral survival.

\subsection{Tile Image Acquisition}
\label{subsec:results_imaging}
This section reports the performance of CGRAS for reliable coral image acquisition during the 2025 spawning event and subsequent grow-out period. We further compare the efficiency of CGRAS with the time a trained marine biologist requires to manually produce equivalent data (\tabref{tab:cgras_operational}). In 2025, CGRAS imaged 183 tiles, with each tile resulting in 24 sub-images, with 1 non-recoverable tile imaging session that was interrupted by a power failure. 

The CCVS processed 160 tiles successfully (87.9\%). 17 failures (9.3\%) were primarily caused by image stitching errors, where insufficient visual features prevented reliable image matching. The remaining 5 failures (2.7\%) resulted from partial occlusion or cropping of the black tile-spacer corners due to incorrect tile placement. These black tile-spacers are necessary for the localization of the tile within the stitched image.

During the 2025 spawning event, manual imaging and counting of 160 tiles was estimated to require 213.5 hours of specialized labor, compared with 22.2 hours using CGRAS, representing a 9.6$\times$ reduction in operational time. At production scale, weekly monitoring of 10,000 tiles over a 12-week cycle would therefore require approximately 120,000 hours of manual labor. At a labor cost of $C$ per hour, this corresponds to $120{,}000C$ per cycle (\eg $\$6.0M$ at $C=\$50$/hour), highlighting the substantial labor and cost savings enabled by CGRAS. These estimates exclude CGRAS capital and maintenance costs and are intended to illustrate the potential scale of operational savings.

\begin{table}[t]
\setlength{\tabcolsep}{3pt}
\vspace*{0.15cm}
\caption{Species-specific performance (mAP50; see~\secref{subsec:metrics} for further details) of the CCVS object detectors. Columns correspond to single-species detectors, while rows correspond to single-species test datasets. Each dataset contains 130 images of live corals from 2024. Best result per column indicated in \textbf{bold}.} 
\label{tab:detection_comparison}
\centering
\scriptsize
\begin{tabularx}{\columnwidth}{@{}l>{\centering\arraybackslash}X>{\centering\arraybackslash}X>{\centering\arraybackslash}X}
\toprule
\textbf{Species} & \textit{Aken} & \textit{Amil} & \textit{Pdae} \\
\midrule
\textit{Aken} & \textbf{96.3} & 91.1 & 64.1   \\
\textit{Amil} & 82.3 & \textbf{96.2} & 60.2  \\
\textit{Pdae} & 50.2 & 55.4  & \textbf{92.3} \\
\arrayrulecolor{black!100}\bottomrule 
\end{tabularx}
\end{table}

\subsection{Coral Detection} 
\label{subsec:results_detection}

\subsubsection{Minimum Data Required}
Given the high cost of manual labeling for high-resolution data, this experiment investigates the minimum number of labeled tile images required to obtain satisfactory performance from the coral detector. This study was conducted on the 2024 dataset and focuses on the coral species \textit{Aken}, for which 100 sub-images were available with an even distribution across the 12 week grow-out period.

In~\figref{fig:detection_results}, we evaluate how the performance of the coral object detector varies with the size of the training set. The model was trained incrementally in batches of 10 tile images, progressively increasing the training set from 10 to 100 tile images. Fixed validation and test sets of 15 tile images each were used to ensure a fair comparison across experiments.

As expected, increasing the number of labeled images improved detection performance at different confidence levels, measured by mAP50. The results show substantial gains up to approximately 30 tile images, where the model achieves an mAP50 above the target performance threshold of 80\%. Beyond this point, improvements are limited, with mAP50 around 85\% only when the full set of 100 tile images is used; however, this ${\approx}5$\% improvement does not justify the annotation cost associated with the additional 70 tile images. Using the full training dataset, the model achieves a precision of 89.1\%, corresponding to 10.9\% false positives, and a recall of 85.9\%, indicating relatively few missed detections, yielding an overall F1-score of 87.5\%. These results are used to guide and balance the labeling effort when introducing new coral species for training new coral species detection models and demonstrate that the model reliably detects coral recruits and provides meaningful counts on grow-out tiles, supporting effective monitoring of the grow-out process.

\begin{figure}[t]
\centering
\subfloat[]{ \includegraphics[width=0.47\columnwidth]{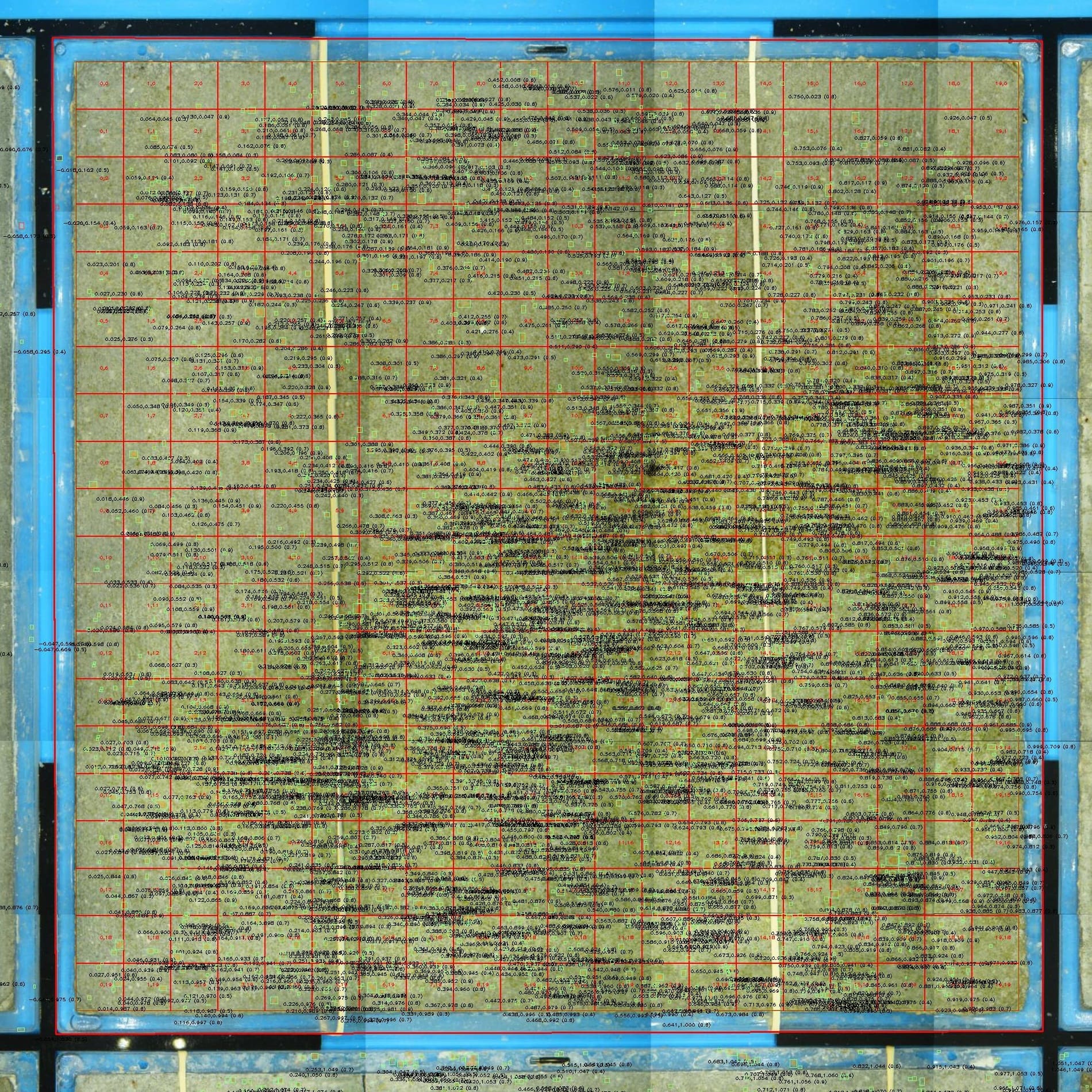} }\hfil
\subfloat[]{ \includegraphics[height=0.465\columnwidth, width=0.47\columnwidth]{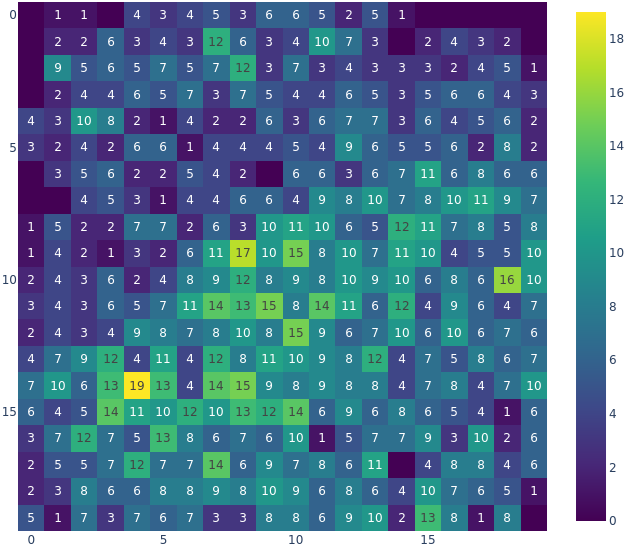} } \caption{Outputs from the CCVS subsystem. (a) Alive corals (indicated by green squares) detected on a stitched tile image, composed of 24 stitched images. The $20\times20$ grid of tabs is overlaid in red. The black corner tile spacers are detected and used to orient the tile within the stitched image. (b) The corresponding coral heatmap (with absolute coral quantities) with respect to the tab grid, where brighter colors represent a greater number of live corals.} 
\label{fig:detections} 
\vspace{-1.0em} 
\end{figure}

\subsubsection{Generalization of Coral Detection Models}
This experiment evaluates the generalization of coral detection models to different coral species. A separate model was trained for each of the three coral species and subsequently evaluated on the test sets of the other species. For this experiment, the 2024 dataset was used as it contained all three coral species. A balanced subset of 130 images of alive corals per species was used, with the results summarized in~\tabref{tab:detection_comparison}.

Strong cross-performance between \textit{Aken} and \textit{Amil} (\figref{fig:coral_examples}), with mAP50 exceeding 82.3\%, is expected given their shared genus (\ie, both \textit{Aken} and \textit{Amil} are part of the genus \textit{Acroporidae}) and similar morphology during early developmental stages. In contrast, \textit{Platygyra daedalea} (\textit{Pdae}) belongs to a different genus and exhibits substantially different morphology and growth patterns, resulting in reduced transfer performance, with mAP50 reaching at most 64.1\%. These results suggest that genus-level detection models (\eg an \textit{Acroporidae} detector instead of specific \textit{Aken}/\textit{Amil} detectors) may be a practical choice for deployment within the CCVS.

\subsection{Coral Density Distribution on Tiles}
\label{subsec:results_heatmaps}

\figref{fig:detections} shows a tile image annotated with detections from the Acropora coral detection model, alongside a visualized heatmap of the alive coral density distribution on a per-tab basis. The heatmap is displayed in a web interface following analysis, enabling operators to quickly identify coral hotspots and better understand how environmental factors, such as tile position and orientation in the grow-out tanks, affect coral health. Furthermore, the CCVS allows the user to select coral tabs for deployment based on the desired density of recruits, which varies based on coral species and deployment objectives. 

\begin{figure}[t]
    \centering
    \includegraphics[width=1.0\columnwidth, trim=0 0 0 1.4cm, clip]{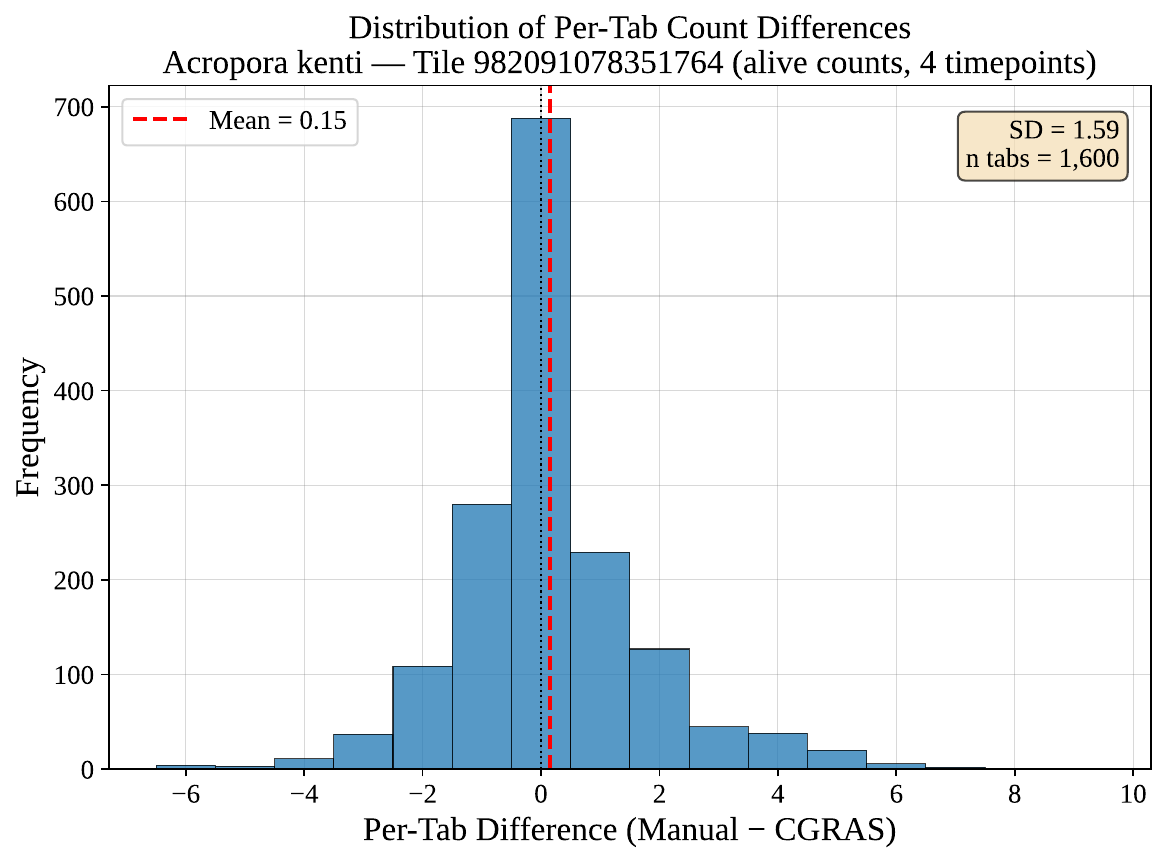}
    \caption{
    Distribution of the differences between \textit{Aken} manual vs CGRAS-based counts per tab. The distribution has a mean of 0.15 miscounted corals per tab, and a standard deviation of 1.59, demonstrating a strong agreement between manual and automated counts. 
    }
    \label{fig:manualcomparison}
\vspace{-1.0em} 
\end{figure}

For validation, manual counts were performed on stitched tile images with the tab grid overlaid, without displaying any model detections to the operator. Validation against manual counts yielded the results in~\figref{fig:manualcomparison}, which shows the distribution of per-tab differences in counts. The majority of per-tab counts are correct with a difference of zero between the manual and our CCVS counts. 

\subsection{Coral Survival Trends}
\begin{figure}
    \centering
    \includegraphics[width=1.0\columnwidth, trim=0 0 0 1.75cm, clip]{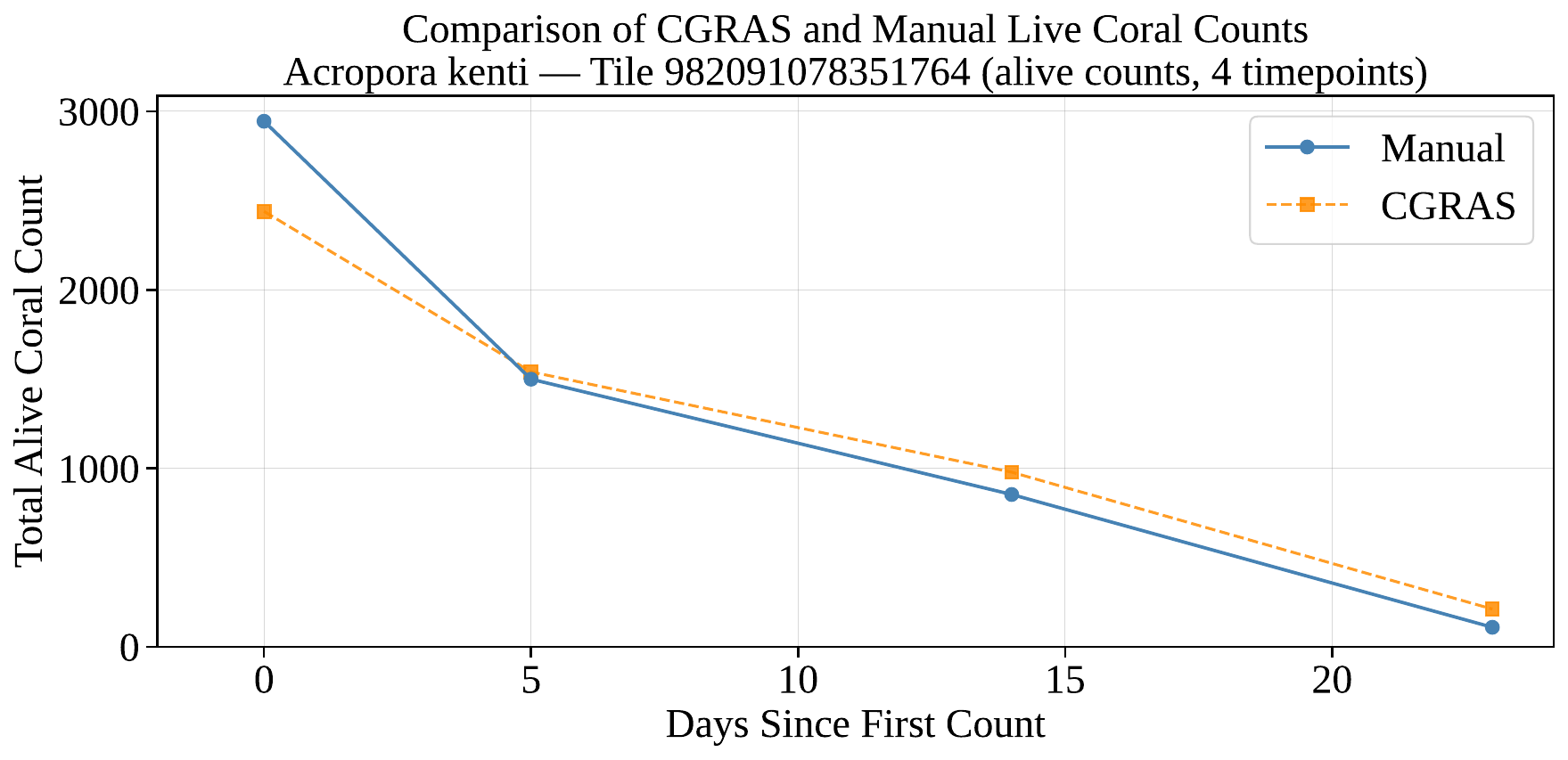}
    \caption{Number of live corals on a tile of \textit{Aken} from 2025, demonstrating the correspondence between the manual counts (blue) and the CGRAS counts (orange).
    }
    \label{fig:heatmap_history}
\vspace{-1.0em} 
\end{figure}

By tracking heatmap outputs across repeated imaging sessions of the same tile, CGRAS generates temporal coral survival trends based on the alive coral detections. An example is shown in \figref{fig:heatmap_history}, where the same tile was imaged and assessed four times over a 23-day period, with each point representing the total number of living corals detected at that assessment.

The comparison between CGRAS and manual counts (\figref{fig:heatmap_history}) resulted in a mean absolute error of 193.3 corals per tile, and differed from manual counts by only 4.4\% on average. While manual counts were used as the reference for evaluation, they may also be subject to human factors such as fatigue and counting variability. Despite these discrepancies, CGRAS provides rapid insights into temporal survival trends. In this example, the estimated number of living corals declined from 2,944 to 109 over 23 days, immediately highlighting substantial mortality and enabling early investigation of potential issues in the grow-out process or aquaculture system. Such automated trend analysis could form the basis of future alert systems or predictive survival models.

\section{Discussion}
\label{sec:discussion}

The proposed system demonstrated strong agreement between automated and manual coral counts, supporting the use of computer vision to reduce the effort required for large-scale coral aquaculture monitoring. The ablation results further showed that useful performance can be achieved with relatively small labeled datasets, highlighting the potential of ML-assisted labeling to accelerate dataset generation.

A key finding of this study is that detector performance depends on the similarity between deployment data and the training distribution. The largest counting discrepancies occurred when coral recruits exhibited visual characteristics that differed substantially from those represented in the training data, leading to confusion with visually similar background objects. This result highlights the importance of fine-tuning models using representative examples of anticipated deployment conditions and suggests that additional training data spanning a wider range of biological states and imaging conditions could improve model robustness. 

The results also revealed limitations in densely populated regions, where closely spaced or merged colonies were sometimes identified as a single coral instance. Such errors can contribute to under-counting and indicate the need for improved approaches for separating individual colonies in crowded scenes. 

Several opportunities for future work remain. The current IAS operates as an open-loop imaging system that relies on accurate calibration and repeatable tile positioning, motivating future investigation into closed-loop approaches and improved optical configurations. While the robotic arm platform provided a flexible research environment, alternative architectures such as protected rail-based systems may offer improved scalability and lower cost for large-scale aquaculture deployments, while portable imaging systems remain important for remote field applications. Within the CCVS, future work will focus on improving image stitching robustness, leveraging unlabeled data through unsupervised or weakly supervised learning, and extending coral detection toward more fine-grained classification across species and developmental stages. Additional directions include incorporating temporal information into growth analysis to better support monitoring and improve processes in coral aquaculture.

\section{Conclusion}
\label{sec:conclusions}
This work presented CGRAS, an integrated robotic imaging and analysis platform that enables consistent, frequent coral survival data that was previously infeasible due to time constraints and labor costs. CGRAS reconciles microscope-scale imaging requirements, marine environment hardware constraints, and high-resolution automated detection within a single field-deployed system. CGRAS operated reliably throughout the 2024 and 2025 coral spawning events, with the IAS enabling large-batch image acquisition with minimal operator intervention. 

The CCVS achieved strong coral detection performance and provided visualization tools for quantifying spatial and temporal survival trends. CGRAS achieved a 96.4\% agreement with manual counts for \textit{Aken} corals, and reduced monitoring effort 9.6-fold compared with manual operation. Critically, the resulting data enables tank conditions and treatment strategies to be correlated with coral survival dynamics at a resolution manual monitoring cannot match. CGRAS demonstrates the potential of automated, scalable monitoring for coral conservation aquaculture and large-scale reef restoration.

While challenges remain in handling crowded scenes and ensuring robust performance across diverse coral appearances, the results demonstrate the potential of computer vision to substantially reduce the effort required for large-scale coral aquaculture monitoring. Future work will focus on improving system robustness and extending the approach to additional coral species and developmental stages. Despite these remaining challenges, CGRAS is already enabling the collection of high-resolution survival data at scales that were previously impractical, providing a powerful new tool to support reef restoration at ecologically meaningful scales.

\bibliographystyle{IEEEtran}
\bibliography{bibliography}

\end{document}